\documentclass[letterpaper]{article} 
\usepackage{aaai23}  
\usepackage{times}  
\usepackage{helvet}  
\usepackage{courier}  
\usepackage[hyphens]{url}  
\usepackage{graphicx} 
\usepackage{natbib}  
\usepackage{caption} 
\usepackage{algorithm}
\usepackage{algorithmic}
\usepackage{amsmath}
\usepackage{amsfonts}
\usepackage{amssymb}
\usepackage{subfig}
\usepackage{booktabs}
\usepackage{multirow}

\usepackage{newfloat}
\usepackage{listings}
\DeclareCaptionStyle{ruled}{labelfont=normalfont,labelsep=colon,strut=off} 
\floatstyle{ruled}
\newfloat{listing}{tb}{lst}{}
\floatname{listing}{Listing}
\newcommand{\image}{I}
\newcommand{\width}{W}

\newcommand{\headnum}{h}
\newcommand{\height}{H}
\newcommand{\patchnumber}{N}
\newcommand{\patch}{t}

\newcommand{\patchind}{n}
\newcommand{\patchsize}{p}

\newcommand{\channelall}{C}
\newcommand{\query}{\mbox{\boldmath $Q$}}
\newcommand{\dquery}{\mbox{\boldmath $\hat Q$}}
\newcommand{\key}{\mbox{\boldmath $K$}}
\newcommand{\val}{\mbox{\boldmath $V$}}
\newcommand{\scorevec}{\mbox{\boldmath $O$}}
\newcommand{\score}{O}

\newcommand{\toksize}{D}
\newcommand{\encoder}{\mbox{\boldmath $	Z$}}
\newcommand{\panel}{\mbox{\boldmath $J$}}
\newcommand{\headsize}{d}

\newcommand{\panelnum}{L}

\newcommand{\dtoken}{\mbox{\boldmath $\hat T$}}
\newcommand{\dencoder}{\mbox{\boldmath $\hat Z$}}
\newcommand{\token}{\mbox{\boldmath $T$}}
\newcommand{\decoder}{\mbox{\boldmath $S$}}
\newcommand{\escore}{\mbox{\boldmath $\hat S $}}

\newcommand{\weight}{\mbox{\boldmath $\cal W$}}

\newcommand{\tbl}[1]{Table \ref{#1}}
\newcommand{\fig}[1]{Fig. \ref{#1}}
\newcommand{\eq}[1]{Eq. \ref{#1}}
\title{Data-Efficient Image Quality Assessment with Attention-Panel Decoder}

\author {
	Guanyi Qin\textsuperscript{\rm 1\thanks{Equal contribution, $\dagger$ Corresponding author.}},
	Runze Hu\textsuperscript{\rm 2\footnotemark[1]},
	Yutao Liu\textsuperscript{\rm 3},
	Xiawu Zheng\textsuperscript{\rm 4,5},
	Haotian Liu\textsuperscript{\rm 1},
	Xiu Li\textsuperscript{\rm 1$\dagger$},
	Yan Zhang\textsuperscript{\rm 5},
}
\affiliations {
	\textsuperscript{\rm 1} Tsinghua Shenzhen International Graduate School, Tsinghua University, Shenzhen 518055, China\\
 \textsuperscript{\rm 2} School of Information and Electronics, Beijing Institute of Technology, Beijing 100086, China\\
	\textsuperscript{\rm 3} School of Computer Science and Technology, Ocean University of China, Qingdao 266100, China\\
	\textsuperscript{\rm 4} Peng Cheng Laboratory, Shenzhen 518066, China\\
	\textsuperscript{\rm 5} Media Analytics and Computing Lab, Department of Artificial Intelligence, School of Informatics, Xiamen University, Xiamen 361005, China\\
	\{qgy21, liu-ht21\}@mails.tsinghua.edu.cn, \{hrzlpk2015, bzhy986\}@gmail.com, liuyutao@ouc.edu.cn, zhengxw01@pcl.ac.cn, li.xiu@sz.tsinghua.edu.cn
}

\begin{document}

\maketitle

\begin{abstract}
Blind Image Quality Assessment (BIQA) is a fundamental task in computer vision, which however remains unresolved due to the complex distortion conditions and diversified image contents. To confront this challenge, we in this paper propose a novel BIQA pipeline based on the Transformer architecture, which achieves an efficient quality-aware feature representation with much fewer data. More specifically, we consider the traditional fine-tuning in BIQA as an interpretation of the pre-trained model. In this way, we further introduce a Transformer decoder to refine the perceptual information of the CLS token from different perspectives. This enables our model to establish the quality-aware feature manifold efficiently while attaining a strong generalization capability. Meanwhile, inspired by the subjective evaluation behaviors of human, we introduce a novel attention panel mechanism, which improves the model performance and reduces the prediction uncertainty simultaneously. The proposed BIQA method maintains a lightweight design with only one layer of the decoder, yet extensive experiments on eight standard BIQA datasets (both synthetic and authentic) demonstrate its superior performance to the state-of-the-art BIQA methods, i.e., achieving the SRCC values of 0.875 (vs. 0.859 in LIVEC) and 0.980 (vs. 0.969 in LIVE). Checkpoints, logs and code will be available at \url{https://github.com/narthchin/DEIQT}.
\end{abstract}

\section{Introduction}

The goal of Image Quality Assessment (IQA) approaches is to automatically evaluate the quality of images in accordance with human subjective judgement. With the increasing growth of computer vision applications, the efficient and reliable IQA model has increased in importance. It is essential to monitor and improve the visual quality of contents and can be also adopted as testing criteria or optimization goals for benchmarking image processing algorithms. 
\begin{figure}[h!] 
	\centering
	\hspace*{-7mm}
        \includegraphics[width=0.80\columnwidth]{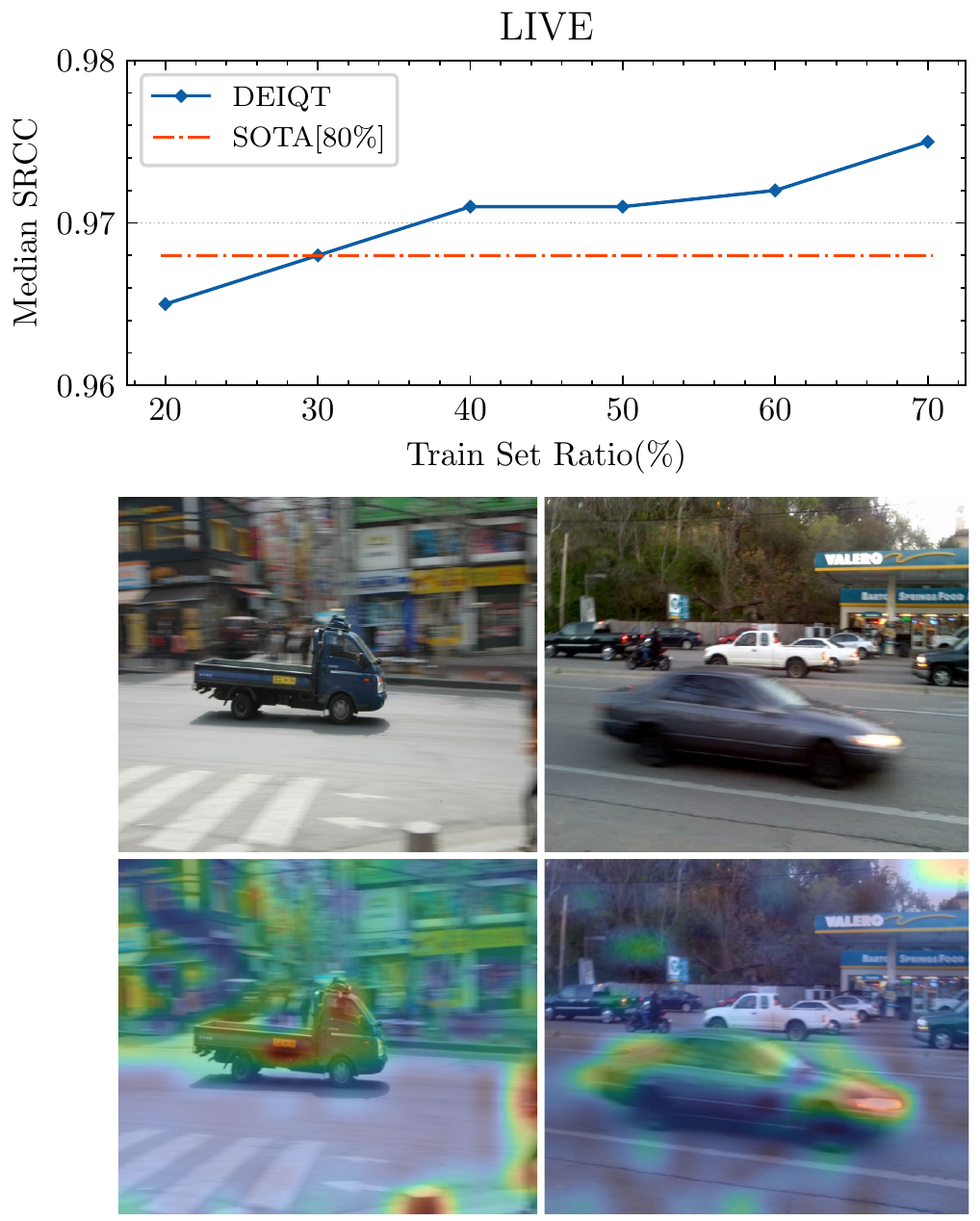}
	
	\caption{Image on top: the performance of the proposed DEIQT varying the amount of training data on the LIVE dataset. SOTA results are obtained from the TReS \cite{golestaneh2022no} using 80\% data. Our method can achieve the SOTA performance with only 30\% data. Images in the medium: the sample images. Images at bottom: Quality attention map from DEIQT. Our model can accurately capture the quality degradation areas of an image. Meanwhile, it ignores the perceptual information that is related to the image recognition yet less important for the quality assessment, i.e., the white cars in the second image.}
	\label{fig:demo}
\end{figure}
Based on the availability of the pristine reference image, IQA can be typically divided into full-reference IQA (FR-IQA) \cite{wang2004image}, reduced-reference IQA (RR-IQA) \cite{soundararajan2011rred}, and no-reference or blind IQA (BIQA) \cite{moorthy2011blind}. 
The applications of FR and RR IQA methods tend to be limited, since reference images are generally unavailable in real-world situations. Correspondingly, the BIQA methods do not require such reference images and thus become more promising yet more challenging.

Current state-of-the-art (SOTA) BIQA methods employ either convolutional neural networks (CNN) or Vision Transformer (ViT) based architectures \cite{dosovitskiy2020image}, which perform an end-to-end optimization of feature engineering and quality regression, simultaneously. The training strategy of BIQA methods generally follows a straightforward pre-training and fine-tuning pipeline. In the pre-training stage, models are trained on a large-scale classification dataset, i.e., ImageNet \cite{deng2009imagenet}. Then, models are fine-tuned on a small-scale BIQA dataset. Nevertheless, the requirements of feature representations for these two stages are not consistent. The pre-training stage concentrates on the global semantic features that are highly related to the image content, whereas the fine-tuning stage needs to consider both global semantics and local details of an image \cite{raghu2021do}. Consequently, the process of fine-tuning still necessitates a substantial amount of data so as to successfully adapt the model awareness from the image content understanding to the image quality. However, due to the labor-intensive characteristics of image annotation, BIQA has high expectations for fitness on low data volumes. Thus, an efficient data-learning strategy, which is capable of constructing an accurate quality-aware feature manifold using a small quantity of data, is desired and has become a beneficial endeavor for computer vision tasks and industrial applications.

To this end, we propose a novel BIQA method that can efficiently characterize the image quality using much fewer data than existing BIQA methods. The proposed BIQA method is based on the Transformer encoder-decoder architecture, herein namely data-efficient image quality transformer (DEIQT). Specifically, we consider that learned features at the pre-training stage are highly related yet more abstract for the BIQA task. In other words, the fine-tuning from the classification task to the BIQA task can be regarded as an interpretation of feature abstractness. Based on this, the classification (CLS) token in the Transformer encoder is an abstract characterization of quality-aware features \cite{cait}. Thus, it may not effectively develop an optimal feature representation for the image quality during the process of fine-tuning. To address this issue, we introduce the Transformer decoder to further decode the CLS token, thereby effectively adapting the token for the BIQA task. 

In particular, we make use of the self-attention and cross-attention operations in the decoder to realize an optimal feature representation for the image quality. The self-attention decodes the aggregated features in the CLS token. It can diminish the significance of those features that are less relevant to the image quality. The resulting outputs of self-attention are handled as the query to the decoder, which is therefore more sensitive to quality-aware image features. The cross-attention performs the interactions between the query and the extracted features from the encoder. It refines the decoder embeddings such that making the extracted features highly related to the image quality. The Transformer decoder brings in an efficient learning property for DEIQT. This not only allows the DEIQT to accurately characterize the image quality using significantly fewer data (\fig{fig:demo}), but also improves the model training efficiency (\fig{fig:dataepoch}). Notably, one layer decoder is adequate to deliver a satisfactory performance for DEIQT (\tbl{tab:abllaynum}), which ensures a lightweight design of our model.

Furthermore, due to the considerable variation in the image contents and distortion conditions, existing BIQA methods generally suffer from a high prediction uncertainty. This hinders the model stability, leading to an inaccurate prediction. To address this issue, we design a novel attention-panel mechanism in the decoder. This mechanism is inspired by the subjective evaluation system, wherein an image is scored by a number of participants and the mean of scores (MOS) is considered the label of this image. During the subjective quality evaluation, opinions of humans on an image differ from person to person. The attention-panel mechanism mimics such human behaviors by randomly initializing and further learning the opinion of each “human” on the image quality. Specifically, it makes use of the cross-attention of decoder to evaluate the image quality from different perspectives and concludes the quality evaluation based on the results from all of these perspectives. The attention-panel mechanism can improve the model stability while introducing almost zero parameters (\tbl{tab:abla}).

In summary, contributions of this paper are the following:
\begin{itemize}
	\item We make the first attempt to develop a BIQA solution based on the complete Transformer encoder-decoder architecture. We employ the CLS token as inputs to the decoder, to enable the proposed DEIQT to extract comprehensive quality-aware features from an image while attaining a high learning efficiency. To the best of our knowledge, we are the first to leverage the Transformer decoder for the IQA task.
	\item Inspired by the human subjective evaluation, we introduce a novel attention-panel mechanism to further improve the performance of DEIQT while reducing the prediction uncertainty. Notably, the attention-panel mechanism introduces almost no parameters to the model.
	\item We verify DEIQT on 8 benchmark IQA datasets involving a wide range of image contents, distortion types and dataset size. DEIQT outperforms other competitors across all these datasets.
\end{itemize}

 \begin{figure*}[] 
	\centering
	\includegraphics[width=0.70\textwidth]{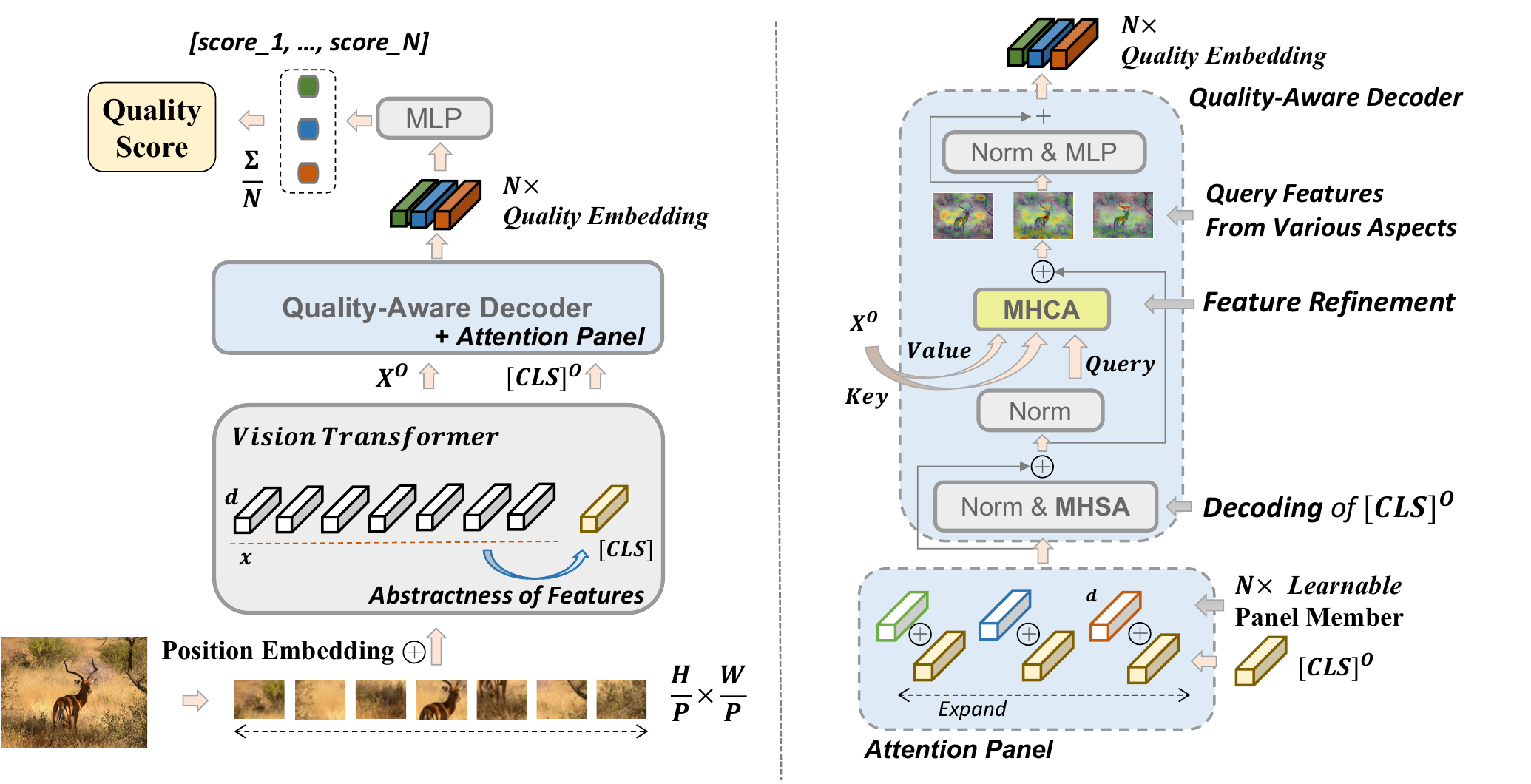}
	\caption{Model overview of the proposed DEIQT}
	\label{fig:arch}
\end{figure*}
\section{Related Work}
\textbf{CNN-based BIQA}. Benefiting from the powerful feature expression ability, CNN-based BIQA methods have gained a great deal of popularity recently \cite{ma2017end, zhang2018blind, su2020blindly, bosse2017deep}. One of the mainstreams of the CNN-based method \cite{kim2016fully} is to integrate the feature learning and regression modeling into a general CNN framework so that developing an accurate and efficient quality representation. Modern CNN-based models \cite{zhang2018blind} also put great efforts into other perspectives of the BIQA challenges, i.e., the limited size of IQA dataset and complex distortion conditions.

In summary, CNN-based methods demonstrate great potential for BIQA tasks, but further efforts are required. Specifically, CNN-based methods usually adopt the image patches \cite{Zhu_2020_CVPR, su2020blindly} as inputs or extract learned features from different layers of CNNs to form a multi-perceptual-scale representation, i.e., the shallow layer for local details and the deeper layer for high-level semantics \cite{hu2021toward,hu2021blind}. The effectiveness of these strategies has been proved, but it introduces non-negligible a computational burden in training and inference\cite{zhang2021perturbed, zhang2022}. Furthermore, due to the inherent locality bias of CNNs, the CNN-based BIQA methods are often constrained by the ineffective characterization of non-local features, notwithstanding the fact that BIQA task depends on both local and non-local image information.

\noindent\textbf{Transformers in BIQA}. Transformers \cite{vaswani2017attention} that were first designed for the natural language processing have raised considerable research interests in the computer vision area. The Vision Transformer (ViT) \cite{dosovitskiy2020image} is one of the most representative works. It performs the classification task using a pure Transformer encoder architecture, and, with modern training strategies, ViT achieves a competing performance against the CNN-based methods \cite{pmlr-v139-touvron21a}. Transformer also demonstrates great potential in dealing with the BIQA task thanks to its strong capability in modelling the non-local dependencies among perceptual features of the image. 
Currently, there are mainly two ways for using Transformer in BIQA: hybrid Transformer \cite{golestaneh2022no, you2021transformer} and pure ViT-based Transformer \cite{ke2021musiq}. The former utilizes CNNs to extract the perceptual features as inputs to the Transformer encoder, whereas the latter directly sends image patches as inputs to the Transformer encoder. 

The Transformer-based BIQA methods have achieved great performance. However, the Transformer in BIQA can be further exploited. Current Transformer-based BIQA methods only involve the Transformer encoder, yet their ability to accurately characterize the image quality is still restricted. The main reason can be attributed to that the extracted features from the encoder are rather abstract in terms of the image quality, making it difficult to model the relations between these features and the quality score. Thus, additional efforts are needed to derive an optimal feature representation for the image quality.

\section{Data-Efficient Image Quality Transformer}

\subsection{Overall Architecture}

To further improve the learning efficiency and capacity of BIQA, we make the first attempt to develop a Transformer encoder-decoder BIQA framework, namely data-efficient image quality transformer (DEIQT). The overall architecture of the proposed DEIQT is illustrated in \fig{fig:arch}. Given an input image, we first obtain the CLS token through the outputs of the Transformer encoder, which acts as the multi-perceptual-level image representation. With the self-attention operation, the CLS token can capture local and non-local dependencies from patch embeddings, thereby preserving comprehensive information for the image quality. The CLS token is then integrated with the attention-panel embeddings via the element-wise addition. A multi-head self-attention block is applied to transform them into queries of the decoder. Each attention-panel embedding absorbs the information from the CLS token, where the cross-attention mechanism in the decoder allows each to learn quality-aware features of an image from a unique perspective. Following this, the transformer decoder outputs the quality embeddings consisting of quality-aware features of an image. Finally, the quality embeddings are sent to a multi-layer perceptron (MLP) head to make several predictions for the image quality. We can obtain one prediction from each embedding of the quality embeddings. The average of these predictions is treated as the final quality score of the image.

\subsection{Perceptual Feature Aggregation in Transformer Encoder}
The self-attention of Transformer encoder aggregates local and non-local information from a sequence of input patches with a minimum inductive bias, which allows it to comprehensively characterize perceptual features of an image. We herein take the advantage of the self-attention to obtain an efficient perceptual representation for the image. Given an input image $ \image \in \mathbb{R}^{\channelall\times \height \times \width} $, we first reshape it into $\patchnumber$ patches as in $\patch_{\patchind} \in  \mathbb{R}^{\patchsize^2 \times \channelall}$, where $\height$ and $\width$ are the height and width of the image, respectively. $\channelall$ is the number of channels and $\patchsize$ indicates the patch size. The total number of patches is calculated as in $ \patchnumber = \frac{\height\width}{\patchsize^2}$. Each patch is then transformed into a $\toksize$-dimension embedding through a linear projection layer. A learnable embedding of CLS token $\token_{\text{CLS}} \in\mathbb{R}^{1 \times \toksize} $ is prepended to the $\patchnumber$ patch embeddings yielding to a total number of $\patchnumber + 1$ embeddings. An additional position embedding is also introduced into these $\patchnumber + 1$ embeddings for preserving the positional information.

Let $\token = \{\token_{\text{CLS}}, \token_{1}, \dots , \token_{\patchnumber} \} \in\mathbb{R}^{ \patchnumber + 1 \times \toksize}$ be the embedding sequence. $\token$ is then fed to the multi-head self-attention (MHSA) block to perform the self-attention operation. The MHSA block contains $\headnum$ heads each with the dimension $\headsize = \frac{\toksize}{\headnum} $. $\token$ is transformed into three groups of matrices as in the query $\query$, key $\key$ and value $\val$ using three different linear projection layers, where  $\query=\{\query_1, ..., \query_{\headnum} \} \in \mathbb{R}^{(\patchnumber+1) \times \toksize}$, $\key=\{\key_1, ..., \key_{\headnum} \} \in \mathbb{R}^{(\patchnumber+1) \times \toksize}$, and  $\val=\{\val_1, ..., \val_{\headnum} \} \in \mathbb{R}^{(\patchnumber+1) \times \toksize}$ for $\query_{\headnum}, \key_{\headnum}, \val_{\headnum}  \in  \mathbb{R}^{(\patchnumber+1) \times \headsize}$.
The output of Transformer encoder $\encoder_{O}$ is formulated as :
\begin{eqnarray}
\begin{split}
	\text{MHSA}\left(\query, \key, \val \right)  = &\text{Cat}(\text{\it Attention}(\query_1, \key_1, \val_1),  \dots, \\
	&\text{~~~~} \text{\it Attention}(\query_{\headnum}, \key_{\headnum}, \val_{\headnum}) ) \weight_{L}\\
	\encoder_{M}  = &  \text{MHA}\left(\query, \key, \val \right) + \token \\
	\encoder_{O}  = & \text{MLP}\left(\text{Norm}(\encoder_{M})\right) + \encoder_{M},
\end{split}
\end{eqnarray}
where  $\weight_{L}$  refers to the weights of the linear projection layer, $ \text{\it Attention}(\query_{\headnum}, \key_{\headnum}, \val_h) = \text{\it softmax}\left(\frac{\query_h {\key_h}^{T}}{\sqrt{d}} \right)\val_h$ and $\text{Norm}(\cdot)$ indicates the layer normalization. $\encoder_{O}$ is denoted as in $\encoder_{O} = \{\encoder_{O}\left[0 \right], ..., \encoder_{O}\left[{\patchnumber} \right] \} \in \mathbb{R}^{(\patchnumber+1) \times \headsize}$.

\subsection{Quality-Aware Decoder}
For the ViT-based BIQA methods, the learned CLS token $\encoder_{O}\left[0 \right]$ is typically considered to contain aggregated perceptual information for the image quality. It will be sent to an MLP head to perform the regression task of quality prediction. However, as explained earlier, $\encoder_{O}\left[0 \right]$  mainly relates to the abstractness of quality-aware features. It is difficult to directly utilize  $\encoder_{O}\left[0 \right]$ to attain an optimal representation for the image quality. To this end, we introduce a quality-aware decoder to further interpret the CLS token, such that making the extracted features more significant to the image quality. 

Let $\dtoken_{\text{CLS}} \in\mathbb{R}^{1 \times \toksize} $ be the CLS token obtained from the output of encoder. $\dtoken_{\text{CLS}}$ is first sent to a MHSA block  to model the dependencies between each element with the remaining elements of the CLS token.  The output of MHSA is followed by the residual connection to generate queries of the transformer decoder, written by
\begin{eqnarray}
	\label{eq:query_decoder}
	\query_{d} = \text{MHSA}\left(\text{Norm}\left(\dtoken_{\text{CLS}}\right) \right) + \dtoken_{\text{CLS}}.
\end{eqnarray}
The role of the MHSA block is to decode the CLS token such that making the produced query more sensitive to the quality-aware features. Following this, we utilize $\dencoder_{O} = \{\encoder_{O}\left[1 \right], ..., \encoder_{O}\left[{\patchnumber} \right] \} \in \mathbb{R}^{\patchnumber\times \headsize} $ as Key and Value of the decoder, denoted by $\key_{d} = \val_{d} = \dencoder_{O}$, where $\dencoder_{O} \cap \encoder_{O} = \dtoken_{\text{CLS}}$.
Then, $\query_{d}, \key_{d}$ and $\val_{d}$ are sent to a multi-head cross-attention (MHCA) block to perform the cross-attention. During this process, we utilize $\query_{d}$ to re-interact with the features of the image patches preserved in the encoder outputs, and thus ensuring the attentional features more significant to the image quality. The output of the cross-attention is written by
\begin{eqnarray}
	\label{eq:cross-att}
	\decoder = \text{MLP}\left(\text{MHCA}(\text{Norm}(\query_{d}), \key_{d}, \val_{d}) + \query_{d} \right),
\end{eqnarray}
where $\decoder$ indicates the refined quality-aware features from the encoder outputs which is more comprehensive and accurate in defining the image quality. Finally, $\decoder$ is fed to an MLP head to derive the final quality score, wherein we minimize the smooth $l_1$ loss to train our network. The quality-aware decoder can significantly improve the learning capacity of the transformer-based BIQA model, and thus enhancing the model performance in terms of prediction accuracy, generalization capability and stability.
\begin{figure}[H] 
	\centering
	\includegraphics[width=0.70\columnwidth]{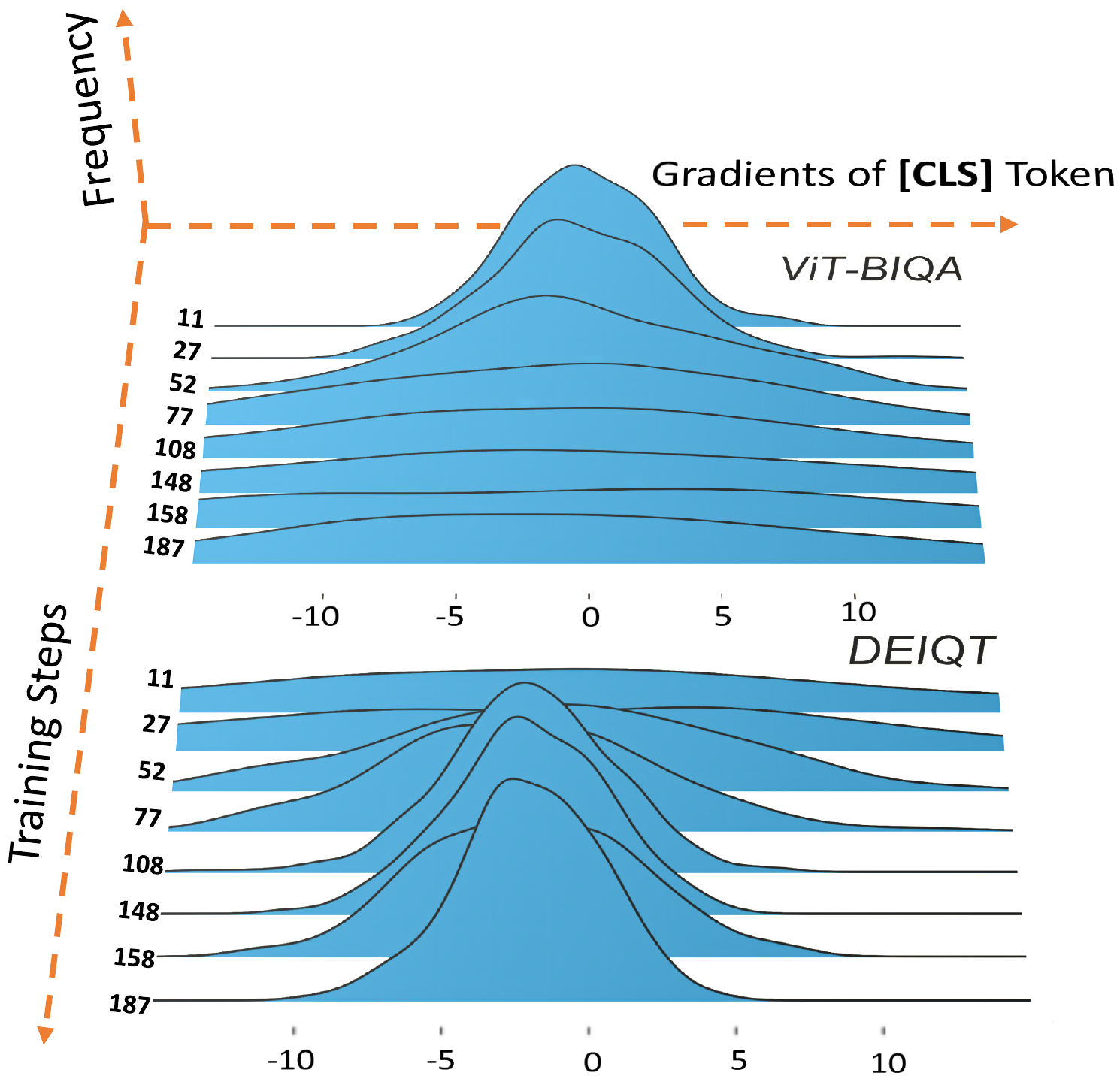}
	\caption{Probability distributions of CLS token Gradients varying the training steps. ViT-BIQA and DEIQT are models without and with the proposed decoder, respectively. By introducing the decoder, variations in the gradients decrease considerably faster than those without the decoder, indicating that the decoder can greatly improve training efficiency.}
	\label{fig:gradcls}
\end{figure}

In \fig{fig:gradcls}, we demonstrate the effectiveness of the quality-aware decoder by investigating the gradients of the CLS token for models with and without the decoder. As observed, without the decoder, the gradients of the CLS token vary significantly throughout the training. This will substantially decrease the training efficiency, and even cause the model to fail to converge. Correspondingly, the designed decoder is capable of reducing such a large variation, thereby ensuring a high training efficiency. It is also worth mentioning that the designed Decoder is combined with the standard ViT encoder in a non-intrusive manner, which not only makes our model compatible with any variants of the ViT encoder but also enables us to directly utilize the weights of other pretrained encoders to increase the training efficiency of our model. More importantly, we empirically show that the designed Decoder with a depth of 1 can effectively achieve a satisfactory performance (\tbl{tab:abllaynum}). This significantly restricts the model size, making it more suitable for practical applications.

\subsection{Attention-Panel Mechanism}

Images captured in the real-world generally involve various contents and complex distortion conditions, resulting in the BIQA models exhibiting a high prediction uncertainty. To mitigate this, we propose an attention-panel mechanism in the Transformer decoder. This mechanism is inspired by the human subjective evaluation, wherein an image is scored by a number of participants and the mean of scores (MOS) is considered the label of the image. During this evaluation process, the personal subjective opinion on an image differs from person to person. The proposed attention-panel mechanism imitates such a situation, in which each panel member represents a participant of the subjection evaluation and judges the image quality from a different perspective. This way, the model can achieve a comprehensive evaluation of the image quality, thus reducing the prediction uncertainty \cite{hu2019statistical}. 

Let $\panelnum$ be the number of panel members. Prior to sending the CLS token to the decoder, we create the attention-panel embeddings as in
$\panel = \{\panel_{1}, \dots, \panel_{\panelnum} \} \in \mathbb{R}^{\panelnum \times \toksize}$. $\panel$ is initialized with random numbers. Then, we expand the CLS token $\panelnum$ times to form the matrix $\dtoken = \{\dtoken_{\text{CLS}}, \dots, \dtoken_{\text{CLS}}\} \in \mathbb{R}^{\panelnum \times \toksize}$. The element-wise summation of $\panel$ and $\dtoken$ is employed as the inputs to the quality-aware decoder. Therefore, the calculation of queries in \eq{eq:query_decoder} is reformulated as
\begin{eqnarray}
	\dquery_{d} = \text{MHSA}\left(\text{Norm}\left(\dtoken_{\text{CLS}} + \panel\right) \right) + \left(\dtoken_{\text{CLS}} + \panel \right).
\end{eqnarray}
The operation of cross-attention is performed in \eq{eq:cross-att} by replacing $\query_{d}$ with $\dquery_{d}$. We obtain the quality embeddings $\escore = \{\escore_1, ... \escore_{\panelnum}   \} \in \mathbb{R}^{\panelnum \times \toksize}$. Finally, $\escore$ is sent to the MLP to derive a vector of scores as in $\scorevec = \{\score_1, \dots, \score_{\panelnum} \}$, which contains $\panelnum$ scores corresponding to $\panelnum$ members. The mean of these $\panelnum$ scores $\displaystyle \frac{\sum_{l=1}^{\panelnum}}{\panelnum}$ is treated as final quality score. 

\begin{figure}[!ht] 
	\centering
	\includegraphics[width=0.90\columnwidth]{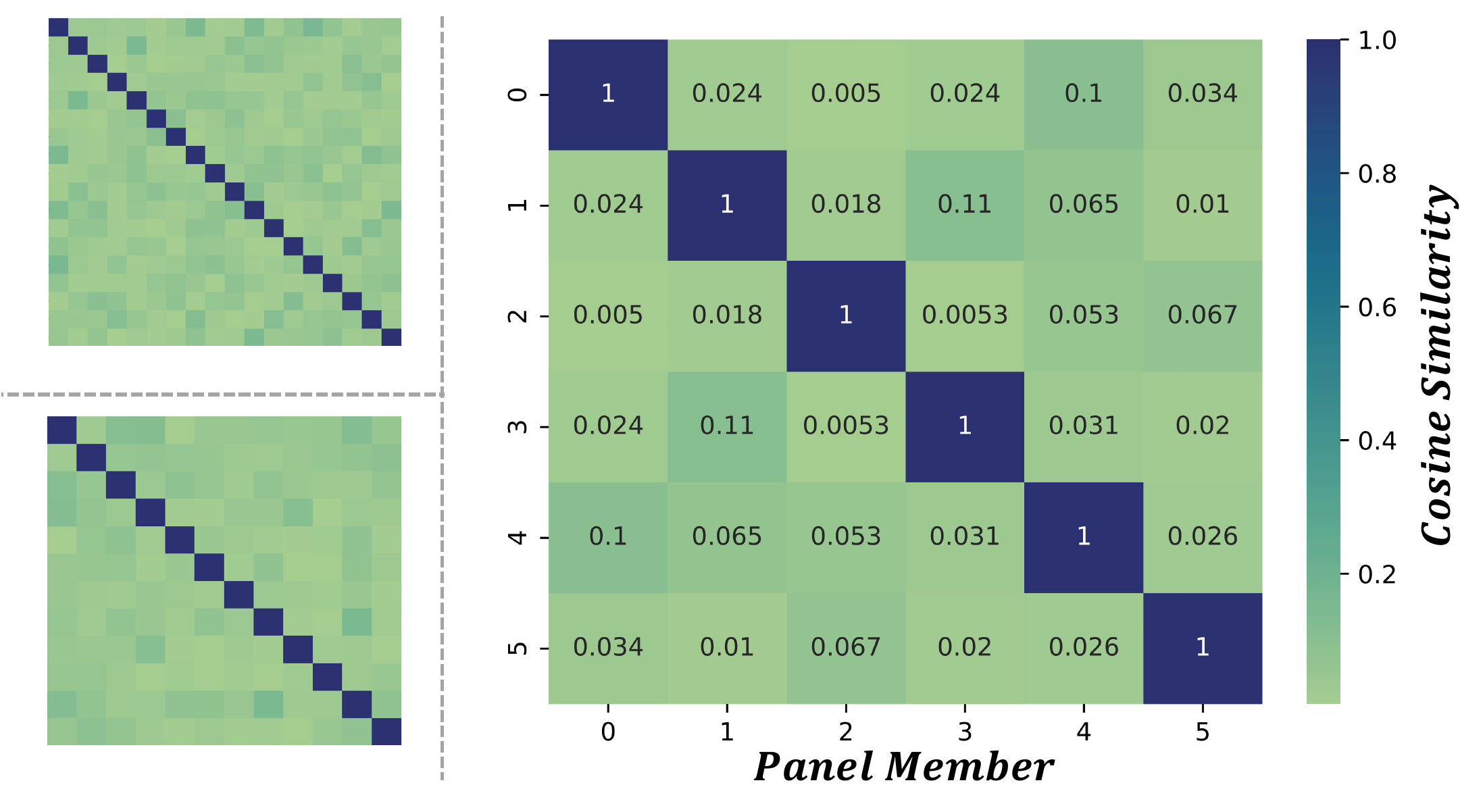}
	\caption{Cosine similarity between characterized perceptual features of panel members. The number of panel members in DEIQT is set to 6, 12, and 18, respectively. The extremely low similarity between two members suggests that each member judges the image quality from a very unique perspective.}
	\label{fig:panel}
\end{figure}

With the attention-panel, DEIQT is capable of characterizing the image quality from different perspectives, thus attaining a comprehensive evaluation. To verify that, we adopt the cosine similarity metric to measure the similarity between the characterized perceptual features from any two panel members. Given an image, we obtain the quality embeddings from three trained DEIQT models with 6, 12 and 18 panel members, respectively. The cosine similarity between every two quality embeddings is computed. The results are reported in \fig{fig:panel}. As observed, the similarity between these panel members is extremely low. Accordingly, the quality-aware features described by each panel member are rather different.
\section{Experiments}
\subsection{Benchmark Datasets and Evaluation Protocols}

We evaluate the performance of the proposed DEIQT on 8 typical BIQA datasets, including 4 synthetic datasets of LIVE \cite{sheikh2006statistical}, CSIQ \cite{larson2010most}, TID2013 \cite{ponomarenko2015image}, KADID \cite{lin2019kadid}, and 4 authentic datasets of LIVEC \cite{ghadiyaram2015massive}, KONIQ \cite{hosu2020koniq}, LIVEFB \cite{ying2020patches}, SPAQ \cite{fang2020perceptual}. Specifically, for the authentic datasets, LIVEC consists of 1162 images captured by different photographers with a wide variety of mobile devices. SPAQ contains 11000 images collected by 66 smartphones. KonIQ-10k is composed of 10073 images selected from public multimedia resources. LIVEFB is the largest-scale authentic dataset (by far) that includes 39810 images. For the synthetic datasets, they contain a small number of pristine images which are synthetically distorted by various distortion types, such as JPEG compression and Gaussian blurring. LIVE and CISQ contain 799 and 866 synthetically distorted images with 5 and 6 distortion types, respectively. TID2013 and KADID consist of 3000 and 10125 synthetically distorted images involving 24 and 25 distortion types, respectively.

In our experiments, two commonly used criteria, Spearman’s rank order correlation coefficient (SRCC) and Pearson’s linear correlation coefficient (PLCC), are adopted to quantify the performance of DEIQT in terms of prediction monotonicity and prediction accuracy, respectively. Both SRCC and PLCC range from 0 to 1. A superior performance should result in the absolute values of SRCC and PLCC close to 1.

\begin{table*}[]
	\centering
	\resizebox{\textwidth}{!}{
		\begin{tabular}{lcccccccc||cccccccc}
			\toprule[1.5pt] & \multicolumn{2}{c}{ LIVE } & \multicolumn{2}{c}{ CSIQ } & \multicolumn{2}{c}{ TID2013 } & \multicolumn{2}{c||}{ KADID } & \multicolumn{2}{c}{ LIVEC } & \multicolumn{2}{c}{ KonIQ } & \multicolumn{2}{c}{ LIVEFB } & \multicolumn{2}{c}{ SPAQ } \\
			\cmidrule{2-17}
			\Large{Method} & PLCC & SRCC & PLCC & SRCC  & PLCC & SRCC & PLCC & SRCC & PLCC & SRCC & PLCC & SRCC & PLCC & SRCC & PLCC & SRCC\\
			\midrule[1pt] 
			DIIVINE & 0.908 & 0.892 & 0.776 & 0.804 & 0.567 & 0.643 & 0.435 & 0.413 & 0.591 & 0.588 & 0.558 & 0.546 & 0.187 & 0.092 & 0.600 & 0.599 \\
			BRISQUE  & 0.944 & 0.929 & 0.748 & 0.812 & 0.571 & 0.626 & 0.567 & 0.528 & 0.629 & 0.629 & 0.685 & 0.681 & 0.341 & 0.303 & 0.817 & 0.809 \\
			ILNIQE  & 0.906 & 0.902 & 0.865 & 0.822 & 0.648 & 0.521 & 0.558 & 0.534 & 0.508 & 0.508 & 0.537 & 0.523 & 0.332 & 0.294 & 0.712 & 0.713 \\
			BIECON & 0.961 & 0.958 & 0.823 & 0.815 & 0.762 & 0.717 & 0.648 & 0.623 & 0.613 & 0.613 & 0.654 & 0.651 & 0.428 & 0.407 & - & - \\
			MEON & 0.955 & 0.951 & 0.864 & 0.852 & 0.824 & 0.808 & 0.691 & 0.604 & 0.710 & 0.697 & 0.628 & 0.611 & 0.394 & 0.365 & - & - \\
			WaDIQaM  & 0.955 & 0.960 & 0.844 & 0.852 & 0.855 & 0.835 & 0.752 & 0.739 & 0.671 & 0.682 & 0.807 & 0.804 & 0.467 & 0.455 & - & - \\
			DBCNN  & \textbf{0.971} & 0.968 & \textbf{0.959} & \textbf{0.946} & 0.865 & 0.816 & 0.856 & 0.851 & 0.869 & 0.851 & 0.884 & 0.875 & 0.551 & 0.545 & 0.915 & 0.911 \\
			TIQA & 0.965 & 0.949 & 0.838 & 0.825 & 0.858 & 0.846 & 0.855 & 0.850 & 0.861 & 0.845 & 0.903 & 0.892 & 0.581 & 0.541 & - & - \\
			MetaIQA & 0.959 & 0.960 & 0.908 & 0.899 & 0.868 & 0.856 & 0.775 & 0.762 & 0.802 & 0.835 & 0.856 & 0.887 & 0.507 & 0.540 & - & - \\
			P2P-BM  & 0.958 & 0.959 & 0.902 & 0.899 & 0.856 & 0.862 & 0.849 & 0.840 & 0.842 & 0.844 & 0.885 & 0.872 & 0.598 & 0.526 & - & - \\
			HyperIQA ({\it 27M}) & 0.966 & 0.962 & 0.942 & 0.923 & 0.858 & 0.840 & 0.845 & 0.852 & \textbf{0.882} & \textbf{0.859} & 0.917 & 0.906 & 0.602 & 0.544 & 0.915 & 0.911 \\
			TReS ({\it 152M})& 0.968 & \textbf{0.969} & 0.942 & 0.922 & \textbf{0.883} & \textbf{0.863} & 0.858 & 0.859 & 0.877 & 0.846 & \textbf{0.928} & 0.915 & 0.625 & 0.554 & - & - \\
			MUSIQ ({\it 27M})& 0.911 & 0.940 & 0.893 & 0.871 & 0.815 & 0.773 & \textbf{0.872} & \textbf{0.875} & 0.746 & 0.702 & \textbf{0.928} & \textbf{0.916} & \textbf{0.661} & \textbf{0.566} & \textbf{0.921} & \textbf{0.918} \\
			\midrule[1pt]
			DEIQT ({\it 24M}) (Ours) & \textbf{0.982} & \textbf{0.980} & \textbf{0.963} & \textbf{0.946} & \textbf{0.908} & \textbf{0.892} & \textbf{0.887} & \textbf{0.889} & \textbf{0.894} & \textbf{0.875} & \textbf{0.934} & \textbf{0.921} & \textbf{0.663} & \textbf{0.571} & \textbf{0.923} & \textbf{0.919}\\
			\bottomrule[1.5pt]
	\end{tabular}}
	\caption{Performance comparison measured by medians of SRCC and PLCC, where bold entries indicate the top two results.}
\label{tab:all}
\end{table*}
\subsection{Implementation Details}
\label{sec:imple}
For DEIQT, we followed the typical training strategy to randomly crop an input image into 10 image patches with a resolution of $224 \times 224$. Each image patch was then reshaped into a sequence of patches with the patch size $\patchsize = 16$, and the dimensions of input tokens $\toksize = 384$. We created the Transformer encoder based on the ViT-S proposed in DeiT III \cite{Touvron2022DeiTIR}. The depth of the encoder was set to 12, and the number of heads $\headnum = 6$. For the Decoder, the depth was set to 1  and the number of panel members $\panelnum = 6$. 

The Encoder of DEIQT was pre-trained on the ImageNet-1K for 400 epochs using the Layer-wise Adaptive Moments optimizer \cite{You2020Large} for Batch training with the batch size 2048. DEIQT was trained for 9 Epochs. The learning rate was set to $2\times 10^{-4}$ with a decay factor of 10 every 3 epochs. The batch size was determined depending on the size of the dataset, i.e., 16 and 128 for the LIVEC and KonIQ, respectively. For each dataset, 80\% images were used for training and the remaining 20\% images were utilized for testing. We repeated this process 10 times to mitigate the performance bias and the medians of SRCC and PLCC were reported.

\subsection{Overall Prediction Performance Comparison}

\tbl{tab:all} reports the comparison results between the proposed DEIQT and 13 state-of-the-art BIQA methods, which include both hand-crafted BIQA methods, such as DIIVINE \cite{saad2012blind}, ILNIQE \cite{ILNIQE} and BRISQUE \cite{mittal2012no}, and deep-learning-based methods, i.e., MUSIQ \cite{ke2021musiq} and MetaIQA \cite{Zhu_2020_CVPR}. As observed across these eight datasets, DEIQT outperforms all other methods. Since images on these 8 datasets span a wide variety of image contents and distortion types, it is very challenging to consistently achieve the leading performance on all of them. Correspondingly, these observations confirm the effectiveness and superiority of DEIQT in characterizing the image quality. 

\subsection{Generalization Capability Validation}
We further evaluate the generalization capability of DEIQT through the cross-datasets validation methodology, where a BIQA model is trained on one dataset, and then tested on the other datasets without any fine-tuning or parameter adaption. The experimental results in terms of the medians of SRCC on five datasets are reported in \tbl{tab:cross}. As observed, DEIQT  achieves the best performance on four of five datasets and reaches the competing performance on the KonIQ dataset. These results manifest the strong generalization capability of DEIQT.
\begin{table}[!ht]
	
	\centering
	\resizebox{0.95\columnwidth}{!}{ 
		\begin{tabular}{lcccccc}
			\toprule[1.5pt] \Large{Training} & \multicolumn{2}{c}{ LIVEFB } & LIVEC & KonIQ & LIVE & CSIQ \\
			\midrule[0.25pt] \Large{Testing} & KonIQ & LIVEC & KonIQ & LIVEC & CSIQ & LIVE \\
			\midrule[1pt]
			DBCNN  & 0.716 & 0.724 & 0.754 & 0.755 & 0.758 & 0.877 \\
			P2P-BM  & 0.755 & 0.738 & 0.740 & 0.770 & 0.712 & - \\
			HyperIQA  & \textbf{0.758} & 0.735 & \textbf{0.772} & 0.785 & 0.744 & 0.926 \\
			TReS  & 0.713 & 0.740 & 0.733 & 0.786 & 0.761 & - \\
			\midrule[1pt]
			DEIQT & 0.733 & \textbf{0.781} & 0.744 & \textbf{0.794} & \textbf{0.781} & \textbf{0.932} \\
			\bottomrule[1.5pt]
		\end{tabular}
	}
	\caption{SRCC on the cross datasets validation. The best performances are highlighted with boldface.}
	\label{tab:cross}
\end{table}

\subsection{Data-Efficient Learning Validation}
One of the key properties of DEIQT is data-efficient learning, which allows our model to achieve a competing performance to state-of-the-art BIQA methods while requiring substantially less training data. Given the costly image annotation and model training, such a property is highly desired for BIQA methods. To further investigate this property, we conduct controlled experiments to train our model by varying the amount of training data from 20\% to 60\% with an interval of 20\%. We repeat the experiment 10 times for each amount of training data and report the medians of SRCC. The testing data remains 20\% images irrespective of the amount of the training data and is completely nonoverlapped with the training data throughout all experiments.

\begin{table}[h!]
	\centering
	\resizebox{\columnwidth}{!}{
		\begin{tabular}{clcccccc}
			\toprule[1.5pt]
			&  & \multicolumn{2}{c}{LIVE} & \multicolumn{2}{c}{LIVEC} & \multicolumn{2}{c}{KonIQ} \\ \cmidrule{3-8} 
			\Large{Mode} & \Large{Methods} & PLCC & SRCC & PLCC & SRCC & PLCC & SRCC \\ \midrule[1pt]
			& ViT-BIQA & 0.828 & 0.894 & 0.641 & 0.622 & 0.855 & 0.825 \\
			\Large{20$\%$} & HyperNet  & 0.950 & 0.951 & 0.809 & 0.776 & 0.873 & 0.869 \\
			& DEIQT & \textbf{0.968} & \textbf{0.965} & \textbf{0.822} & \textbf{0.792} & \textbf{0.908} & \textbf{0.888} \\ \midrule[0.25pt]
			& ViT-BIQA & 0.847 & 0.903 & 0.714 & 0.684 & 0.901 & 0.880 \\
			\Large{40$\%$} & HyperNet  & 0.961 & 0.959 & 0.849 & 0.832 & 0.908 & 0.892 \\
			& DEIQT & \textbf{0.973} & \textbf{0.971} & \textbf{0.855} & \textbf{0.838} & \textbf{0.922} & \textbf{0.903} \\ \midrule[0.25pt]
			& ViT-BIQA & 0.856 & 0.915 & 0.739 & 0.705 & 0.916 & 0.903 \\
			\Large{60$\%$} & HyperNet  & 0.963 & 0.960 & 0.862 & 0.843 & 0.914 & 0.901 \\
			& DEIQT & \textbf{0.974} & \textbf{0.972} & \textbf{0.877} & \textbf{0.848} & \textbf{0.931} & \textbf{0.914} \\ \bottomrule[1.5pt]
		\end{tabular}
	}
	\caption{Data-efficient learning validation with the training set containing 20\%, 40\% and 60\% images.  Bold entries indicate the best performance.}
	\label{tab:eff}
\end{table}
The experimental results are detailed in \tbl{tab:eff}. On the synthetic datasets, even with 20\% images, DEIQT has reached a competing performance to the second best BIQA method in \tbl{tab:all}. When the training data contains 40\% images, DEIQT outperforms all other BIQA methods. In other words, DEIQT utilizes only half of the training data and achieves the best performance on the synthetic datasets. For authentic datasets, DEIQT is capable of achieving the competing performance with 60\% images, which is still much more efficient than other BIQA methods.
\begin{figure}[h!] 
	\centering
	\hspace*{-3mm}
	\includegraphics[width=0.95\columnwidth]{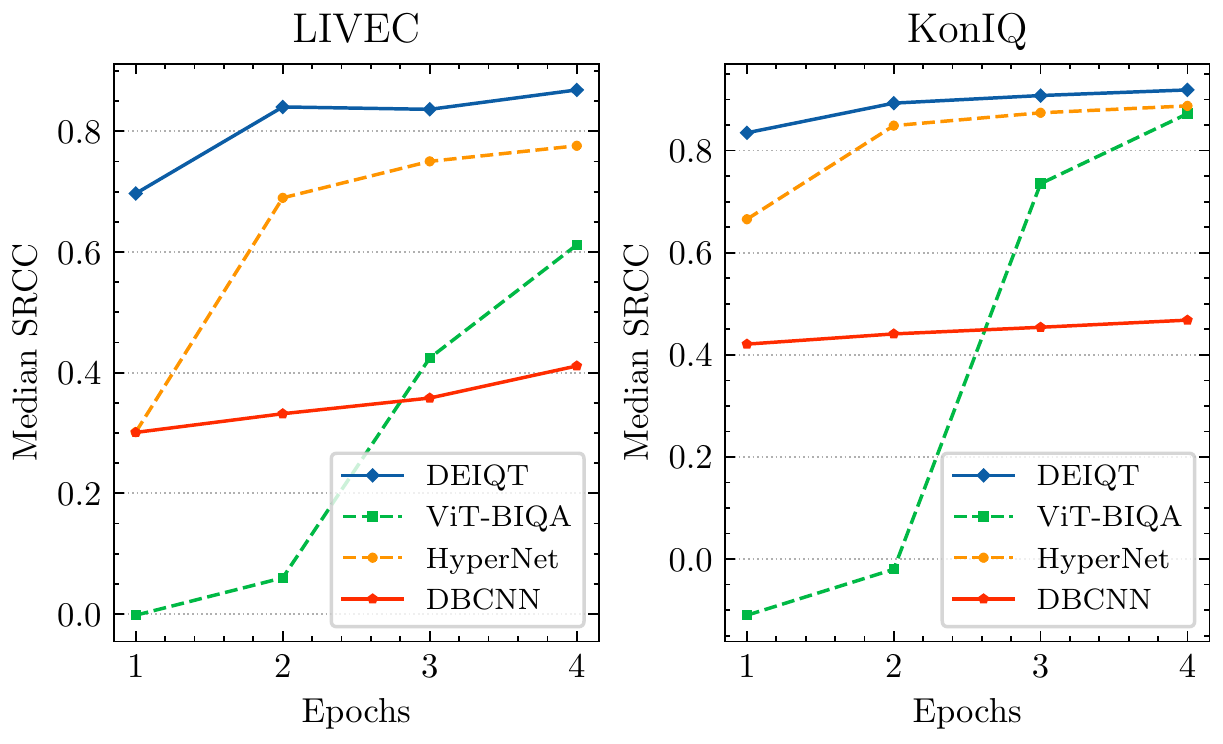}
	\caption{Median SRCC versus Epochs on the LIVEC and KonIQ testing datasets.}
	\label{fig:dataepoch}
\end{figure}

In addition to the required training data, we further evaluate the training efficiency  of DEIQT which is also an important indicator for data-efficient learning. \fig{fig:dataepoch} illustrates the medians of SRCC as the number of epochs increases on the testing set of LIVEC and KonIQ. ViT-BIQA directly utilizes the extracted features of the CLS token to predict the image quality. As shown in \fig{fig:dataepoch}, DEIQT converges significantly faster than other methods, where it reaches a satisfactory performance in only two epochs. As a comparison, ViT-BIQA exhibits a slow convergence rate, especially on the small-scale dataset LIVEC. These observations vividly demonstrate that DEIQT can efficiently implement the domain adaptation from the pre-training of classification tasks to the fine-tuning of IQA tasks, thereby greatly improving the training efficiency.

\subsection{Ablation Study}
DEIQT is composed of three essential components, including the ViT encoder, quality-aware decoder, and the attention-panel mechanism. We conduct the ablation experiments to examine the individual contribution of each component. \tbl{tab:abla} shows the experimental results on the LIVEC and KonIQ datasets. The ViT in \tbl{tab:abla} refers to the DEIQT without the quality-aware decoder and the attention-panel. It is equivalent to the ViT-BIQA in \fig{fig:dataepoch}. AP/6 indicates the attention-panel (AP) with 6 panel members. ViT + AP/6 skips the decoder and sends the inputs of the DEIQT decoder to an MLP head to make the prediction. Decoder($\text{R}^{*}$) and Decoder(CLS) mean that random numbers or CLS token are utilized as inputs of the decoder, respectively. The proposed DEIQT consists of ViT, Decoder(CLS) and AP/6.
\begin{table}[h!]
	\centering
	\resizebox{\columnwidth}{!}{ \Large{
		\begin{tabular}{lccccc}
			\toprule[1.5pt]
			& & \multicolumn{2}{c}{LIVEC} & \multicolumn{2}{c}{KonIQ} \\ \cmidrule{3-6} 
			\LARGE{Module} & \LARGE{\#Params} & PLCC & SRCC & PLCC & SRCC \\ \midrule[1pt]
			ViT & \multirow{2}{*}{22M} &0.770 & 0.714 & 0.919 & 0.908 \\
			std & & $\pm$0.045 & $\pm$0.039 & $\pm$0.011 & $\pm$0.011 \\ \midrule[0.25pt]
			ViT + AP/6 & \multirow{2}{*}{22M} & 0.782 & 0.720 & 0.924 & 0.913 \\
			std & & $\pm$0.033 & $\pm$0.030 & $\pm$0.010 & $\pm$0.008 \\ \midrule[0.25pt]
			ViT + Decoder($\text{R}^{*}$) & \multirow{2}{*}{24M} & 0.871 & 0.842 & 0.927 & 0.916 \\
			std & & $\pm$0.018 & $\pm$0.024 & $\pm$0.007 & $\pm$0.006 \\ \midrule[0.25pt]
			ViT + Decoder(CLS) & \multirow{2}{*}{24M} & 0.881 & 0.863 & 0.931 & 0.918 \\
			std & & $\pm$0.018 & $\pm$0.019 & $\pm$0.005 & $\pm$0.007 \\ \midrule[0.25pt]
			DEIQT & \multirow{2}{*}{24M} & \textbf{0.894} & \textbf{0.875} & \textbf{0.934} & \textbf{0.921} \\
			std & & $\pm$0.014 & $\pm$0.017 & $\pm$0.003 & $\pm$0.004 \\ \bottomrule[1.5pt]
		\end{tabular}}
	}
	\caption{Ablation experiments on LIVEC and KonIQ datasets. Bold entries indicate the best performance.}
\label{tab:abla}
\end{table}

From \tbl{tab:abla}, we observe that both the quality-aware decoder and the attention-panel mechanism are highly effective in characterizing the image quality, and thus contributing to the overall performance of DEIQT. In particular, the proposed quality-aware decoder significantly improves the model performance in terms of accuracy and stability, whereas the attention-panel contributes less than the decoder. This is reasonable considering that the number of parameters introduced by the decoder is substantially higher than those introduced by the attention-panel. The operations involved in the decoder are also much more sophisticated. Nevertheless, the attention-panel allows our model to attain improved performance with negligible additional expense.

Finally, we carry out the experiment to investigate the effects of the decoder depth on the DEIQT. The experimental results are listed in \tbl{tab:abllaynum}. As can be seen that DEIQT is insensitive to the depth of decoder. When the number of layers of decoder increases, the performance of DEIQT remains almost unchanged on these two datasets. Thus, we set the number of layers of decoder to 1 to maintain a lightweight design for our model. 
\begin{table}[h!]
	\centering
	\resizebox{.85\columnwidth}{!}{ \Large  
		\begin{tabular}{cccccc}
			\toprule[1.5pt]  & & \multicolumn{2}{c}{ LIVEC } & \multicolumn{2}{c}{ KonIQ } \\
			\cmidrule{3-6} 
			\LARGE{Layer Nums} & \LARGE{\#Params} & PLCC & SRCC & PLCC & SRCC \\
			\midrule[1pt] 1 & 24M &0.894 & 0.875 & 0.934 & 0.921 \\
			2 & 26M &0.890 & 0.871 & 0.933 & 0.919 \\
			4 & 31M &\textbf{0.895} & \textbf{0.877} & \textbf{0.936} & \textbf{0.922} \\
			8 & 40M &\textbf{0.895} & 0.873 & 0.933 & 0.918 \\
			\bottomrule[1.5pt]
		\end{tabular}
	}
	\caption{The effects of the layer numbers of the decoder on the DEIQT. Bold entries indicate the best results.}
	\label{tab:abllaynum}
\end{table}
\section{Conclusion}
In this paper, we present a data-efficient image quality transformer (DEIQT), which can accurately characterize the image quality using much less data. In particular, we regard the CLS token as the abstractness of quality-aware features and adapt it to the queries of the dedicatedly designed decoder. Then, we leverage the cross-attention mechanism to decouple the quality-aware features from the encoder outputs. Furthermore, inspired by the human behaviors of the subjective evaluation, we offer a novel attention-panel mechanism to mitigate the prediction uncertainty while introducing almost no additional parameters. Experiments on eight standard datasets demonstrate the superiority of DEIQT in terms of prediction accuracy, training efficiency, and generalization capability.

\section{Acknowledgments}
This research was partly supported by the National Key R\&D Program of China (Grant No. 2020AAA0108303), and in part by National Natural Science Foundation of China under Grant 62201538, Natural Science Foundation of Shandong Province under grant ZR2022QF006, and in part by China National Postdoctoral Program for Innovative Talents (BX20220392), China Postdoctoral Science Foundation (2022M711729)

\bibliography{aaai23}

\end{document}